\documentclass{article}

    \PassOptionsToPackage{numbers, compress}{natbib}

\usepackage[final]{neurips_2021}
\usepackage{graphicx}

\usepackage[utf8]{inputenc} 
\usepackage[T1]{fontenc}    
\usepackage{url}            
\usepackage{booktabs}       
\usepackage{amsfonts}       
\usepackage{nicefrac}       
\usepackage{microtype}      
\usepackage[linesnumbered, ruled, vlined]{algorithm2e}
\usepackage{algorithmic}
\usepackage{pifont}
\usepackage[table,dvipsnames]{xcolor}
\title{Sparsity-based Feature Selection for Anomalous Subgroup Discovery}

\author{%
 Girmaw Abebe Tadesse \\
 IBM Research Africa \\
 \texttt{girmaw.abebe.tadesse@ibm.com } \\
 \And
 William Ogallo\\
 IBM Research Africa \\
 \texttt{william.ogallo@ibm.com } \\
 \And 
 Catherine Wanjiru \\
 Carnegie Mellon University Africa\\
 \texttt{ccatheri@andrew.cmu.edu} \\
 \And
 Charles Wachira \\
 IBM Research Africa \\
 \texttt{charles.wachira1@ibm.com} \\
 \And
 Isaiah Onando Mulang' \\
 IBM Research Africa\\
 \texttt{mulang.onando@ibm.com}\\
 \And
 Vibha Anand\\
 IBM Research - Cambridge\\
 \texttt{anand@us.ibm.com}\\
\And
 Aisha Walcott-Bryant \\
 IBM Research Africa \\
 \texttt{alwalcott@ke.ibm.com}\\
\And
 Skyler Speakman \\
 IBM Research Africa \\
 \texttt{skyler@ke.ibm.com}
}

\begin{document}

\maketitle

\begin{abstract}
Anomalous pattern detection aims to identify instances where deviation from normalcy is evident, and is widely applicable across domains. Multiple anomalous detection techniques have been proposed in the state of the art. However, there is a common lack of a principled and scalable feature selection method for efficient discovery. Existing feature selection techniques are often conducted by optimizing the performance of  prediction  outcomes  rather than its systemic deviations from the expected. In this paper, we proposed a sparsity-based automated feature selection (SAFS) framework, which encodes systemic outcome deviations via the sparsity of feature-driven odds ratios. SAFS is a model-agnostic approach with usability across different discovery techniques.  SAFS achieves more than $3\times$ reduction in computation time while maintaining detection performance when validated on publicly available critical care dataset. SAFS also results in a superior performance when compared against multiple baselines for feature selection.
\end{abstract}

\section{Introduction}
\label{sec:intro}

Detection of anomalous samples (aka outlier or novelty detection) is a field of active research that aims to identify observations (a subgroup of samples) in a given data that deviate from some concept of normality~\cite{ruff2021unifying}. Its application is crucial across different domains that include  healthcare~\cite{ogallo2021detection,kim2021out,zhao2019deep}, cybersecurity~\cite{xin2018machine},  insurance and finance sectors~\cite{zheng2018generative}, and industrial monitoring~\cite{hundman2018detecting}.
 The challenges associated with anomalous detection primarily span three themes. First, the lack of representative examples of anomalous cases results in a significant imbalance that limits the choice of detection approaches. Second, the variations among in-distribution samples might be equivalent (even worse) compared to the deviation of anomalous samples, resulting in a rise in type-I and type-II errors. Finally, the scope of anomalousness could be too wide to model, with extreme variations of anomalous cases. 
 
 Nevertheless, a plethora of methods has been proposed for anomalous detection. The methods could be mainly categorised into \textit{reconstruction}, \textit{classification} and \textit{probabilistic} groups~\cite{ruff2021unifying}.  The well-known principal component analysis and autoencoders are examples of reconstruction-based methods that, first, transform the data (e.g., to a latent space) so that anomalousness could be detected from failing to reconstruct the data back from the transformed data~\cite{hawkins2002outlier}. Classification-based approaches, particularly one-class classification is often employed due to the lack of examples representing anomalous cases~\cite{tax2002one,khan2014one}.  Furthermore, the traditional probabilistic models have also been used to identify anomalous samples using estimation of the normal data probability distribution, e.g., Gaussian mixture models~\cite{roberts1994probabilistic} and Mahalanobis distance evaluation~\cite{laurikkala2000informal}. Moreover, there are purely distance-based  methods, such as k-nearest neighbourhood~\cite{gu2019statistical}, that do not require a prior training phase nor data transformations. Of note is that most existing methods infer anomalousness by exploiting individual sample characteristics rather than group-based characteristics. To this end, researchers have developed the Multi-dimensional subset scanning (MDSS), a method that aims to identify subsets of anomalous samples by exploiting group-level characteristics~\cite{mcfowland2018efficient,cintas2021pattern,tadesse3897703principled}.

A significant gap in the state-of-the-art anomalous subgroup discovery concerns the lack of a principled and scalable approach to select input features prior to the discovery of anomalousness. Most discovery techniques require manual selection of features (e.g., using domain experts) or use the whole input space, often resulting in inefficient discovery characterized by long computational time (due to exponentially growing possible combinations of feature values),  less interpretable anomalous characterization without considerable regularisation efforts, and a higher likelihood of a search output being in a local optimum due to the very high dimensional input space. Automated feature selection steps could be employed to solve the above problem by selecting useful features for the subsequent discovery step. However, existing feature selection techniques mainly optimize over higher outcome prediction performance of a trained model~\cite{wanjiru2021automated,molina_2002_feature,miao_2016_a}, and hence they are limited for encode systemic outcome deviations among subsets of the data. Moreover, model training results in computational overhead, and the feature selection output is prone to  model hyper-parameters, class imbalance and underfitting or overfitting . 

In this paper, we proposed a sparsity-based automated feature selection (SAFS) framework, which is \textit{model-agnostic} as it does not require training a particular model. SAFS encodes systemic outcome deviations using the sparsity of the feature-driven odds ratios. The proposed feature selection framework is simple and generalizable as it could be applied as a simple pre-processing step to any anomalous discovery technique for tabular data formats. Specifically, contributions of SAFS are three-fold: 1) SAFS significantly reduces the search space and consequently, the amount of time required to complete the search; 2) SAFS also decreases the number of optimization steps necessary to approximate global optima; and 3) SAFS improves the interpretation of the identified anomalous subgroup as it discards less relevant features early. We validated SAFS on the publicly available  MIMIC-III (Medical Information Mart for Intensive Care) dataset\cite{johnson2016mimic}.
The results show that SAFS achieves similar anomalous subgroup discovery using just half of the features selected while providing more than $3\times$ reduction in computational time. 
Furthermore, we demonstrate that SAFS results in superior detection performance when compared with the state-of-the-art feature selection algorithms including Filters~\cite{molina_2002_feature,vergara2014review}, Wrappers~\cite{miao_2016_a}, and Embedded techniques that require training of  tree-based models~\cite{wanjiru2021automated}, such as XGBoost~\cite{chen2016xgboost}  and Catboost~\cite{hancock2020catboost}.

\section{Proposed framework}
\label{sec:format}
\begin{figure}
    \centering
        \caption{SAFS framework for automatically selecting important features for anomalous subgroup discovery.}\label{fig:overview}
    \includegraphics[width=1\linewidth]{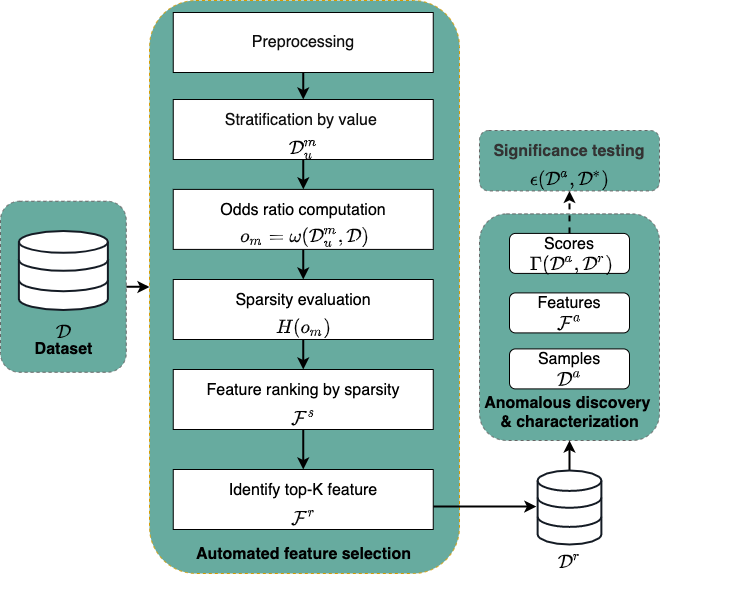}

\end{figure}

The proposed framework is shown in Fig.~\ref{fig:overview} and it contains two main components: \textit{automated feature selection (SAFS)} and \textit{anomalous discovery and characterization}. The automatic feature selection exploits the sparsity of odd ratios computed per each feature value. The anomalous discovery step scans across all possible combinations of the values of the selected features and characterizes those identified to be divergent from the expectation (or normalcy). Each component of the framework is described below in detail. 

\subsection{Problem Formulation}
Let $\mathcal{D}  = \{(x_i,y_i) | i = 1,2,\cdots, N\}$ denotes a dataset containing $N$ samples, and each sample $x_i$ is characterised by a set of $M$ discretized features  $\mathcal{F}=[f_1,f_2,\cdots,f_m, \cdots, f_M]$ and $y_i$ represents the outcome label. Note that each feature $f_m$ has $C_m$ unique values, $\hat{f}_m =\{\hat{f}_m^u\}_{u=1}^{C_m}$.  The proposed automated feature selection process is defined as $\mathcal{R}(\cdot)$ that takes $\mathcal{D}$ as input and  provides $\mathcal{D}^r$ represented with a top $K$ features, i.e., $\mathcal{D}^r=R(\mathcal{D},K) = \{(x^r_i,y_i) | i = 1,2,\cdots, N\}$ and $x^r_i$ is represented by  $\mathcal{F}^r =\{f^r_1,f^r_2,\cdots,f^r_k,\cdots,f^r_K\}$, where $K \leq M$.  Then the anomalous subgroup discovery, $S(\cdot)$, takes $\mathcal{D}^r$  as input and identifies the anomalous subgroup ($\mathcal{D}^a$) represented by the set of anomalous features $\mathcal{F}^a=\{f^a_z\}_{z=1}^Z$, $Z \leq K \leq M$. The overall anomalous feature description is described as the  logical (AND and OR) combinations of anomalous feature values as 
$\hat{\mathcal{F}}^a = \bigcap\limits_{z=1}^Z (\bigcup\limits_{h=1}^{H_z} \hat{f^{a}_{zh}}) $, where $\hat{f^{a}_{zh}}$ represents the $h^{th}$ value of the $f^{a}_z$ and $H_z < C_z$. Note that $\mathcal{F}^a \subseteq \mathcal{F}^r \subseteq  \mathcal{F}$. The anomalous subgroup $\hat{\mathcal{D}}^a$ contains samples from $\mathcal{D}^r$ whose feature values are characterized by $\hat{\mathcal{F}}^a$, i.e.,  $\mathcal{D}^a=\{(x^a_j,y^a_j) | j=1,2,\cdots, P\}$, where $P < N$.
The anomalousness of the identified subgroup is evaluated based on the anomalous score, $\Gamma(\mathcal{D}^a,\mathcal{D}^r)$.

\subsection{Automated Feature Selection}
 The sparsity-based automated feature selection (SAFS) component  in Fig.~\ref{fig:overview} is tasked with selecting the top $K$ features from a given $M$-dimensional feature space that are more useful for the follow up anomalous subgroup discovery. Most state-of-the-art automated anomalous group discovery techniques for tabular data formats search for any possible combination of feature values in the data, which grows exponentially with the addition of a feature. To this end, SAFS employs sparsity of the odds ratios of feature values to rank and select features. A highly-ranked feature for anomalous discovery is assumed to have high sparsity across the odds ratios of its feature values.  Given a feature $f_m$ with $C_m$ unique values, we manually stratify $\mathcal{D}$ per each feature value $\hat{f}_m^u \in \hat{f}_m$, i.e., $\mathcal{D}_{m}^u =\mathcal{D} | \hat{f}_m == \hat{f}_m^u$. The mean of the outcome in the stratified $\mathcal{D}_{m}^u$ is computed as $\mu_m^u =\frac{\sum_{j=1}^{N_m^u} y_j}{N_m^u}$, where $N_{m}^u$ is the number of samples in $\mathcal{D}_m^u$. Similarly, the global average of the outcome is computed as $\mu_g =\frac{\sum_{i=1}^N y_i}{N}$, where $N$ is the total number of samples in $\mathcal{D}$. Thus, ratios of the odds of the outcome in $\mathcal{D}_m^u$  and  in $\mathcal{D}$ is computed as:
 \begin{equation}
 o_m^u=\frac{\mu_m^u/(1-\mu_m^u)}{\mu_{g}/(1-\mu_{g})}\\
 \end{equation}
To compute the sparsity of the odd ratios in $o_m$, we use the Hoyer sparsity metric~\cite{hoyer2004non} that was proven to satisfy key requirements of sparsity~\cite{hurley2009comparing} as follows:
  \begin{equation}
 \eta_m  =(\sqrt{C_m} -\frac{\sum_{u=1}^{C_{m}}o_m^{u}}{\sum_{u=1}^{C_m}{o_m^u}^2})(\sqrt{C_m} - 1)^{-1}
\end{equation}
The summary of the steps for sparsity-based feature selection is shown in Algorithm~\ref{alg:algo_featureselection}.

\begin{algorithm}[t]
\caption{Pseudo-code for automated feature selection based on sparsity of odds ratios}
\label{alg:algo_featureselection}

\SetKwFunction{IdentifyUniqueValues}{IdentifyUniqueValues}
\SetKwFunction{ZerosArray}{ZerosArray}
\SetKwFunction{SortDescendingIndices}{SortDescendingIndices}
\SetKwFunction{TopK}{TopK}
\SetKwFunction{OddsRatio}{OddsRatio}
\SetKwFunction{MeasureSparsity}{MeasureSparsity}
\SetKwFunction{Stratification}{Stratification}

\SetKwInOut{Input}{input}
\SetKwInOut{Output}{output}
\Input{Dataset: $\mathcal{D}$  $= \{(x_i,y_i) | i = 1,2,\cdots, N\}$,\\ Set of features: $\mathcal{F}$ $=[f_1,f_2,\cdots,f_m, \cdots, f_M]$,\\ Required number of features: $K$.}
\Output{Set of selected features: $\mathcal{F}^r$} 
\BlankLine
$\eta$ $\leftarrow$ \ZerosArray($M$)\;
\For{$f_m$ in $\mathcal{F}$}{
$\hat{f}_m$ $\leftarrow$ \IdentifyUniqueValues($f_m$)\;
$C_m$ $\leftarrow$ $|\hat{f}_m|$ \;
$o_{m}$ $\leftarrow$ \ZerosArray($C_m$)\;
\For{$u$ $\leftarrow$ $1 \KwTo C_m$}{
 $\mathcal{D}^u_m$ $\leftarrow$ \Stratification($\mathcal{D}$,$\hat{f}_m^u$)\;
 $o_m^u$ $\leftarrow$ \OddsRatio($\mathcal{D}^u_m$,$\mathcal{D}$)\;
 }

 $\eta_m$ $\leftarrow$ \MeasureSparsity($o_m$)\;
 }
 
 $I$ $\leftarrow$ \SortDescendingIndices($\eta$)\;
 $\mathcal{F}^s$ $\leftarrow$ $\mathcal{F}[I]$\;
 $\mathcal{F}^r$ $\leftarrow$ \TopK($\mathcal{F}^s, K$)\;

\Return $\mathcal{F}^r$
\end{algorithm}

\subsection{Anomalous Discovery and Characterization}
We employ Multi-Dimensional Subset Scanning(MDSS)~\cite{mcfowland2018efficient,cintas2020detecting} from the anomalous pattern detection literature in order to identify significantly divergent subset of samples. Characterization of the identified samples includes quantifying the anomalousness score, the analysis of the anomalous features and their values, the time elapsed to identify them, and the statistical significance of these findings. 
MDSS could be posed as a search problem over possible subsets in a multi-dimensional array to identify systematic deviation between observation (i.e., $y_i$) and expectation of the outcomes, which could be set differently for variants of MDSS. In the simple automatic stratification setting, the expectation is the global outcome average in $\mathcal{D}^r$, i.e., $\mu_g$.
The deviation between the expectation and observation is evaluated by maximizing a Bernoulli likelihood ratio scoring statistic, $\Gamma(\cdot)$. The null hypothesis assumes that the likelihood of the outcome in each sample $x_i^r \in \mathcal{D}^r$ or subgroup is similar to the expected ($\mu_g$), i.e., $H_0: odds(y_i)=\frac{\mu_g}{1-\mu_g}$; while the alternative hypothesis assumes a constant multiplicative increase in the outcome odds for the anomalous subgroup, $H_1: odds(y_i)=q\frac{\mu_g}{1-\mu_g}$ where $q\neq1$ ($q>1$ for extremely over observed subgroup; and $0<q<1$ for extremely under observed subgroup). The  anomalous scoring function for a subgroup ($\mathcal{D}^s$) with reference $\mathcal{D}^r$ is formulated as, $\Gamma(\mathcal{D}^s,\mathcal{D}^r)$ and computed as:
 \begin{equation}
\Gamma(\mathcal{D}^s,\mathcal{D}^r) = \max_q log(q)\sum_{i\in S} y_i - N_S * log(1-\mu_g + q\mu_g),
 \end{equation}
 where $N_S$ is the number of samples in $\mathcal{D}^s$.
Consequently, subsets in which average of outcome different from $\mu_g$ will have higher scores.  Subset identification is iterated until convergence to a local maximum is found, and the global maximum is subsequently optimized using multiple random restarts.  The subset ($\mathcal{D}^s$) with the highest score becomes anomalous subset $\mathcal{D}^a$ and it is characterized by its score of anomalousness, $\Gamma(\mathcal{D}^a,\mathcal{D}^r)$ and a combination of feature values $\hat{\mathcal{F}}^a$ describing the identified $\mathcal{D}^a$.

\subsection{Significance Testing}
The statistical significance of SAFS is evaluated using a randomisation testing. The null hypothesis  suggests  $\Gamma(\mathcal{D}^a,\mathcal{D}^r)$ is not significantly different from a set of $\Gamma(\mathcal{D}^*,\mathcal{D}^r)$, where $\mathcal{D}^*$ represents the anomalous subset obtained from $\mathcal{D}^{r*}$, obtained  by randomly selecting $K$ features from $\mathcal{D}$.  This experiment is performed iteratively $\sigma=100$ times resulting $\delta =\{\Gamma(\mathcal{D}^*_t,\mathcal{D}^r)\}_{t=1}^\sigma$.
We compute the empirical p-value as $\frac{(\xi+1)}{(\sigma+1)}$ where $\xi$ is the number of scores  in $\delta$ that are greater than or equal to the actual score $\Gamma(\mathcal{D}^a,\mathcal{D}^r)$.

\section{Experiments}

\subsection{Dataset and Experimental Setup}
We used the Medical Information Mart for Intensive Care (MIMIC-III) dataset\cite{johnson2016mimic} to validate the proposed framework. 
We selected a study cohort of adult patients (16 years or older) who were admitted to the ICU for the first time, where the length of stay was greater than a day, and with no hospital readmissions, no surgical cases, and having at least one chart events. The final cohort consisted of $N=18,761$ patients. We constructed $M=41$ features based on observations made on the first 24 hours of ICU admission.
We defined the target outcome as a binary indicator variable $y_i$ such that  $y_i=1$ for patients who died within 28 days of the onset of their ICU admission, and $y_i=0$ otherwise. For the automatic selection step, we set top K to different values $\{0.1 \times l \times 41\}_{l=1}^{10}$ resulting $K\in [4,8,12,16,20,25,29,33,37,41]$.

\subsection{Results and Discussion}
%

    

\begin{figure}[t]
    \centering
        \caption{Anomalous subgroup scores (green) and elapsed scanning time (red) across top $K$ features selected by SAFS.}
        \label{fig:score_time}
    \includegraphics[width=0.6\linewidth]{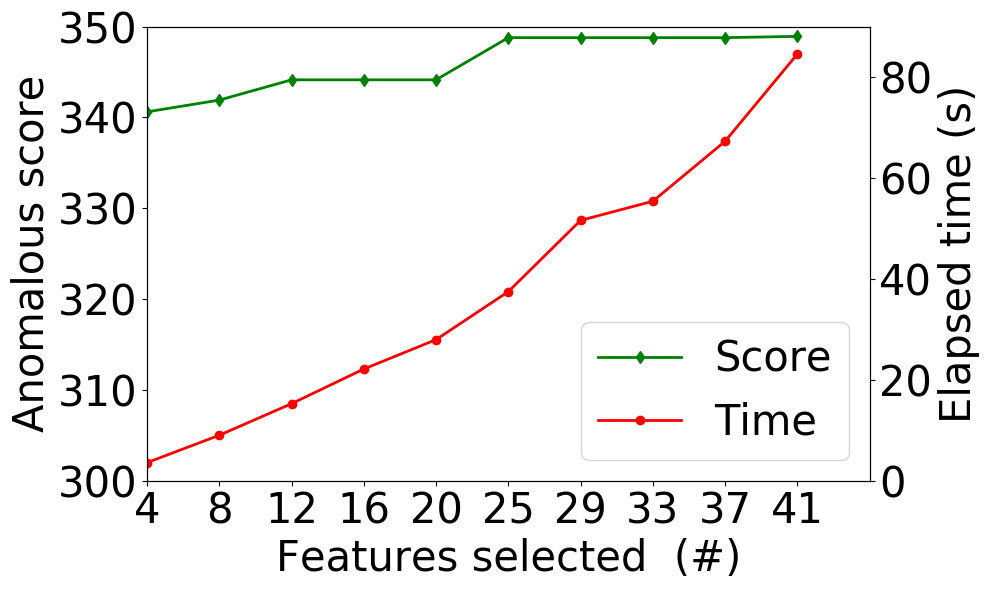}
\end{figure}

%
We selected different top $K$ features using SAFS and apply MDSS to identify anomalous subset characterized by more observations of deaths compared to the global average.  Figure~\ref{fig:score_time} illustrates the anomalous score and the elapsed time to complete the scanning across the top $K$ selected features. These results show that it is possible to achieve comparable anomalous scores by scanning over the top $20$ features (half of the original) identified by SAFS, with more than  $3\times$ reduction in elapsed scanning time.

\begin{table}[]
    \centering
    \caption{Subpopulation size, odds ratio, and significance (empirical p-value) of the most anomalous subgroup when scanning over the  top $K$ features identified by SAFS.}
\resizebox{0.5\linewidth}{!}{
\begin{tabular}{lcclc}
\hline
 Top K & Size &  Odds Ratio ($95\%$ CI) &  P-Value \\ \hline
\rowcolor[HTML]{EFEFEF}
 4 &           4312 &        2.48 (2.3, 2.67) &      $<0.01$ \\
 8 &           2811 &        3.00 (2.75, 3.26) &      $<0.01$ \\
\rowcolor[HTML]{EFEFEF} 12 &           4383 &        2.48 (2.31, 2.67) &      $<0.01$ \\
16 &           4383 &        2.48 (2.31, 2.67) &      $<0.01$ \\
\rowcolor[HTML]{EFEFEF} 20 &           4383 &        2.48 (2.31, 2.67) &      $<0.01$ \\
25 &           3078 &        2.93 (2.7, 3.18) &      $<0.01$ \\
\rowcolor[HTML]{EFEFEF} 29 &           3078 &        2.93 (2.7, 3.18) &      $<0.01$ \\
33 &           3078 &        2.93 (2.7, 3.18) &      $<0.01$ \\
\rowcolor[HTML]{EFEFEF} 37 &           3078 &        2.93 (2.7, 3.18) &      $<0.01$ \\
41 &           4218 &        2.54 (2.36, 2.73) &      $<0.01$ \\ \hline

\end{tabular}
    \label{tab:odds_topks}
    }
\end{table}

\begin{figure}[t]
    \centering
        \caption{Overlapping of anomalous features detected under different top K values.}\label{fig:anom_features}
    \includegraphics[width=0.5\linewidth]{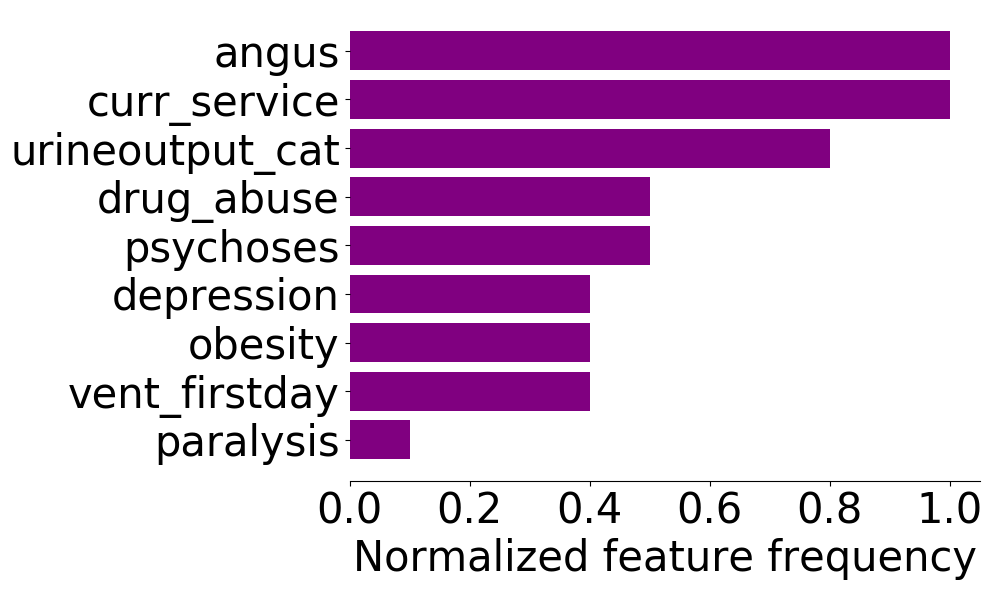}

\end{figure}

In addition to the anomalous score, we also evaluated the consistency of the anomalous group across different top $K$ values as shown in Table~\ref{tab:odds_topks} and Fig.~\ref{fig:anom_features}. 
The results demonstrate that consistently competitive performance is achieved across different top $K$ values without a loss of performance in anomalous discovery. Fig.~\ref{fig:anom_features} shows that consistent features are identified across these $K$ values. E.g., features \textit{angus} and \textit{curr\_service} represented the anomalous group in all top $K$ values, and \textit{urine output} appeared in eight  cases out of $10$ different $K$ values.   Moreover, these most frequently occurring features are ranked higher during the feature selection step, validating the effectiveness of SAFS in selecting features that would be useful for anomalous pattern discovery.





Table~\ref{tab:baseline_comparison} shows the comparison of SAFS with multiple state-of-the-art feature selection approaches including a filter method using mutual information gain ~\cite{molina_2002_feature,vergara2014review}, a wrapper method using recursive feature elimination~\cite{guyon2002gene}, and embedded methods using XGBoost~\cite{chen2016xgboost}  and Catboost~\cite{hancock2020catboost} and committee vote based on the average of feature importance from XGBoost and Catboost. 
The results demonstrate  the superior performance of SAFS to achieve highest anomalous scores, particularly with $K\leq29$ features. All methods become competitive for larger $K$ values. 

\begin{table}[]
    \centering
    \caption{Comparison of the proposed SAFS to existing feature selection methods for different top $K$ values.  SAFS is shown to outperform others with its highest anomalous scores for  $K\leq25$ features. The methods become competitive afterwards. Committee: average of XGB and CatB. }
\resizebox{0.8\linewidth}{!}{
\begin{tabular}{r|rr|rrr|r}
\hline
&&&  \multicolumn{3}{c|}{Embedded methods} &    \\
 K &  Filter~\cite{molina_2002_feature} &  Wrapper~\cite{miao_2016_a} &    XGB~\cite{chen2016xgboost} &   CatB~\cite{hancock2020catboost} &  Committee 
 &   SAFS \\
\hline
     \rowcolor[HTML]{EFEFEF} 4 &              337.19 &  311.27 &   337.19 &    339.23 &     337.19 &    \textbf{340.61 }\\
     8 &              340.61 &  311.27 &   337.19 &    340.61 &     340.61 &    \textbf{341.90 }\\
    \rowcolor[HTML]{EFEFEF} 12 &              340.61 &  311.27 &   340.61 &    340.61 &     340.61 &   \textbf{ 344.13} \\
    16 &              340.61 &  316.09 &   340.61 &    \textbf{344.13} &     340.61 &    \textbf{344.13} \\
    \rowcolor[HTML]{EFEFEF} 20 &              340.61 &  340.86 &   340.61 &    \textbf{344.13} &     \textbf{344.13} &    \textbf{344.13} \\ 

    25 &              340.61 &                         346.64 &   348.53 &    348.45 &     348.53 &    \textbf{348.77} \\ \hline\hline
    \rowcolor[HTML]{EFEFEF} 29 &              344.59 &                         346.64 &   3\textbf{48.77} &    \textbf{348.77} &     \textbf{348.77} &    \textbf{348.77} \\
    33 &              \textbf{348.77} &                         346.64 &   \textbf{348.77} &    \textbf{348.77} &     \textbf{348.77 }&    \textbf{348.77 }\\
    \rowcolor[HTML]{EFEFEF} 37 &              348.45 &                         \textbf{348.91} &   \textbf{348.91} &    \textbf{348.91} &    \textbf{ 348.91} &    348.77 \\
    41 &             \textbf{ 348.91} &                        \textbf{ 348.91} &   \textbf{348.91} &    \textbf{ 348.91} &    \textbf{348.91} &   \textbf{ 348.91} \\
\hline
\end{tabular}
    \label{tab:baseline_comparison}
    }
\end{table}

\section{Conclusion and Future work}
We proposed a sparsity-based automated feature selection (SAFS) framework for anomalous subgroup discovery that aimed to significantly reduce the search space thereby the amount of time required to complete the discovery; to reduce the the number of optimization steps to approximate global optima; and to improve the interpretation of the identified anomalous subgroups.  SAFS is \textit{model-agnostic} with no need of training a model and can be employed as a pre-processing step for most anomalous pattern detection techniques.  SAFS uses the feature-driven deviation of outcome likelihood via the sparsity of the odds ratios to encode systemic deviations. We validated  the framework on a publicly available  MIMIC-III dataset, and results showed that SAFS outperformed multiple baseline methods, and achieved more than $3\times$ reduction in computational time but with competitive detection performance using just half of the features. Future work aims to extend SAFS to select layers and nodes in deep learning frameworks that employ activations and reconstruction errors to identify anomalous patterns in other modalities. In addition, similar framework of feature evaluation could be employed to infer further insights in post-discovery analysis~\cite{mulang2021post}.




\bibliographystyle{unsrt}
\bibliography{neurips_2021}

\begin{thebibliography}{10}

\bibitem{ruff2021unifying}
Lukas Ruff, Jacob~R Kauffmann, Robert~A Vandermeulen, Gr{\'e}goire Montavon,
  Wojciech Samek, Marius Kloft, Thomas~G Dietterich, and Klaus-Robert
  M{\"u}ller.
\newblock A unifying review of deep and shallow anomaly detection.
\newblock {\em Proceedings of the IEEE}, 2021.

\bibitem{ogallo2021detection}
William Ogallo, Girmaw~Abebe Tadesse, Skyler Speakman, and Aisha
  Walcott-Bryant.
\newblock Detection of anomalous patterns associated with the impact of
  medications on 30-day hospital readmission rates in diabetes care.
\newblock In {\em AMIA Annual Symposium Proceedings}, volume 2021, page 495.
  American Medical Informatics Association, 2021.

\bibitem{kim2021out}
Hannah Kim, Girmaw~Abebe Tadesse, Celia Cintas, Skyler Speakman, and Kush
  Varshney.
\newblock Out-of-distribution detection in dermatology using input perturbation
  and subset scanning.
\newblock {\em arXiv preprint arXiv:2105.11160}, 2021.

\bibitem{zhao2019deep}
Rui Zhao, Ruqiang Yan, Zhenghua Chen, Kezhi Mao, Peng Wang, and Robert~X Gao.
\newblock Deep learning and its applications to machine health monitoring.
\newblock {\em Mechanical Systems and Signal Processing}, 115:213--237, 2019.

\bibitem{xin2018machine}
Yang Xin, Lingshuang Kong, Zhi Liu, Yuling Chen, Yanmiao Li, Hongliang Zhu,
  Mingcheng Gao, Haixia Hou, and Chunhua Wang.
\newblock Machine learning and deep learning methods for cybersecurity.
\newblock {\em IEEE Access}, 6:35365--35381, 2018.

\bibitem{zheng2018generative}
Yu-Jun Zheng, Xiao-Han Zhou, Wei-Guo Sheng, Yu~Xue, and Sheng-Yong Chen.
\newblock Generative adversarial network based telecom fraud detection at the
  receiving bank.
\newblock {\em Neural Networks}, 102:78--86, 2018.

\bibitem{hundman2018detecting}
Kyle Hundman, Valentino Constantinou, Christopher Laporte, Ian Colwell, and Tom
  Soderstrom.
\newblock Detecting spacecraft anomalies using lstms and nonparametric dynamic
  thresholding.
\newblock In {\em Proceedings of ACM SIGKDD International Conference on
  Knowledge Discovery \& Data Mining}, pages 387--395, 2018.

\bibitem{hawkins2002outlier}
Simon Hawkins, Hongxing He, Graham Williams, and Rohan Baxter.
\newblock Outlier detection using replicator neural networks.
\newblock In {\em Proceedings of International Conference on Data Warehousing
  and Knowledge Discovery}, pages 170--180, 2002.

\bibitem{tax2002one}
David Martinus~Johannes Tax.
\newblock One-class classification: Concept learning in the absence of
  counter-examples.
\newblock 2002.

\bibitem{khan2014one}
Shehroz~S Khan and Michael~G Madden.
\newblock One-class classification: taxonomy of study and review of techniques.
\newblock {\em The Knowledge Engineering Review}, 29(3):345--374, 2014.

\bibitem{roberts1994probabilistic}
Stephen Roberts and Lionel Tarassenko.
\newblock A probabilistic resource allocating network for novelty detection.
\newblock {\em Neural Computation}, 6(2):270--284, 1994.

\bibitem{laurikkala2000informal}
Jorma Laurikkala, Martti Juhola, Erna Kentala, N~Lavrac, S~Miksch, and
  B~Kavsek.
\newblock Informal identification of outliers in medical data.
\newblock In {\em Fifth International Workshop on Intelligent Data Analysis in
  Medicine and Pharmacology}, volume~1, pages 20--24, 2000.

\bibitem{gu2019statistical}
Xiaoyi Gu, Leman Akoglu, and Alessandro Rinaldo.
\newblock Statistical analysis of nearest neighbor methods for anomaly
  detection.
\newblock {\em arXiv preprint arXiv:1907.03813}, 2019.

\bibitem{mcfowland2018efficient}
Edward McFowland~III, Sriram Somanchi, and Daniel~B Neill.
\newblock Efficient discovery of heterogeneous treatment effects in randomized
  experiments via anomalous pattern detection.
\newblock {\em arXiv preprint arXiv:1803.09159}, 2018.

\bibitem{cintas2021pattern}
Celia Cintas, Skyler Speakman, Girmaw~Abebe Tadesse, Victor Akinwande, Edward
  McFowland~III, and Komminist Weldemariam.
\newblock Pattern detection in the activation space for identifying synthesized
  content.
\newblock {\em arXiv preprint arXiv:2105.12479}, 2021.

\bibitem{tadesse3897703principled}
Girmaw~Abebe Tadesse, Megan~Marx Delaney, Victor Akinwande, William Ogallo,
  Claire-Helene Mershon, Katherine~EA Semrau, and Skyler Speakman.
\newblock Principled subpopulation analysis of the betterbirth study and the
  impact of who's safe childbirth checklist intervention.
\newblock {\em Available at SSRN 3897703}.

\bibitem{wanjiru2021automated}
Catherine Wanjiru, William Ogallo, Girmaw~Abebe Tadesse, Charles Wachira,
  Isaiah~Onando Mulang, and Aisha Walcott-Bryant.
\newblock Automated supervised feature selection for differentiated patterns of
  care.
\newblock {\em arXiv preprint arXiv:2111.03495}, 2021.

\bibitem{molina_2002_feature}
L.C. Molina, L.~Belanche, and A.~Nebot.
\newblock Feature selection algorithms: a survey and experimental evaluation.
\newblock {\em Proceedings of IEEE International Conference on Data Mining},
  2002.

\bibitem{miao_2016_a}
Jianyu Miao and Lingfeng Niu.
\newblock A survey on feature selection.
\newblock 91:919--926, 2016.

\bibitem{johnson2016mimic}
Alistair~EW Johnson, Tom~J Pollard, Lu~Shen, H~Lehman Li-Wei, Mengling Feng,
  Mohammad Ghassemi, Benjamin Moody, Peter Szolovits, Leo~Anthony Celi, and
  Roger~G Mark.
\newblock Mimic-iii, a freely accessible critical care database.
\newblock {\em Scientific Data}, 3(1):1--9, 2016.

\bibitem{vergara2014review}
Jorge~R Vergara and Pablo~A Est{\'e}vez.
\newblock A review of feature selection methods based on mutual information.
\newblock {\em Neural Computing and Applications}, 24(1):175--186, 2014.

\bibitem{chen2016xgboost}
Tianqi Chen and Carlos Guestrin.
\newblock Xgboost: A scalable tree boosting system.
\newblock In {\em Proceedings of ACM SIGKDD International Conference on
  Knowledge Discovery and Data Mining}, pages 785--794, 2016.

\bibitem{hancock2020catboost}
John~T Hancock and Taghi~M Khoshgoftaar.
\newblock Catboost for big data: an interdisciplinary review.
\newblock {\em Journal of Big Data}, 7(1):1--45, 2020.

\bibitem{hoyer2004non}
Patrik~O Hoyer.
\newblock Non-negative matrix factorization with sparseness constraints.
\newblock {\em Journal of Machine Learning Research}, 5(9), 2004.

\bibitem{hurley2009comparing}
Niall Hurley and Scott Rickard.
\newblock Comparing measures of sparsity.
\newblock {\em IEEE Transactions on Information Theory}, 55(10):4723--4741,
  2009.

\bibitem{cintas2020detecting}
Celia Cintas, Skyler Speakman, Victor Akinwande, William Ogallo, Komminist
  Weldemariam, Srihari Sridharan, and Edward McFowland.
\newblock Detecting adversarial attacks via subset scanning of autoencoder
  activations and reconstruction error.
\newblock In {\em International Joint Conference on Artificial Intelligence},
  pages 876--882, 2020.

\bibitem{guyon2002gene}
Isabelle Guyon, Jason Weston, Stephen Barnhill, and Vladimir Vapnik.
\newblock Gene selection for cancer classification using support vector
  machines.
\newblock {\em Machine Learning}, 46(1):389--422, 2002.

\bibitem{mulang2021post}
Isaiah~Onando Mulang, William Ogallo, Girmaw~Abebe Tadesse, and Aisha
  Walcott-Bryant.
\newblock Post-discovery analysis of anomalous subsets.
\newblock {\em arXiv preprint arXiv:2111.14622}, 2021.

\end{thebibliography}

\end{document}